\documentclass[11pt,a4paper]{article}
\usepackage[hyperref]{acl2021}
\usepackage{graphicx}
\usepackage{array}
\usepackage{times}
\usepackage{amsmath}
\usepackage{latexsym}
\usepackage{booktabs}

\usepackage{microtype}

\aclfinalcopy 
\def\aclpaperid{867} 

\title{Sequence to General Tree: Knowledge-Guided \\Geometry Word Problem Solving}

\author{Shih-hung Tsai, Chao-Chun Liang, Hsin-Min Wang, Keh-Yih Su \\
  Institute of Information Science, Academia Sinica, Taiwan \\
  \texttt{\{doublebite,ccliang,whm,kysu\}@iis.sinica.edu.tw} \\}

\date{}

\begin{document}
\maketitle
\begin{abstract}
With the recent advancements in deep learning, neural solvers have gained promising results in solving math word problems. However, these SOTA solvers only generate binary expression trees that contain basic arithmetic operators and do not explicitly use the math formulas. As a result, the expression trees they produce are lengthy and uninterpretable because they need to use multiple operators and constants to represent one single formula. In this paper, we propose sequence-to-general tree (S2G) that learns to generate interpretable and executable \textit{operation trees} where the nodes can be formulas with an arbitrary number of arguments. With nodes now allowed to be formulas, S2G can learn to incorporate mathematical domain knowledge into problem-solving, making the results more interpretable. 
Experiments show that S2G can achieve a better performance against strong baselines on problems that require domain knowledge.\footnote{Data and code are available at the GitHub repository: \href{https://github.com/doublebite/Sequence-to-General-tree/}{https://github.com/doublebite/Sequence-to-General-tree/} }

\end{abstract}

\section{Introduction}


Math word problem (MWP) solving is a special subfield of question answering. It requires machine solvers to read the problem text, understand it, and then compose the numbers and operators into a meaningful equation (as shown in Table \ref{tb:mwp}). This process, even for the simplest problem in elementary school, demands language understanding and numerical reasoning capabilities, making this task a long-standing challenge in AI \cite{bobrow1964natural,zhang2019gap}.

\begin{table}[h!]
\begin{center}
\resizebox{\columnwidth}{!}{%
\begin{tabular}{l l}
 \toprule
\textbf{Problem:} The outer radius and the inner radius \\
of a circular annulus are 5m and 3m repsectively. \\
Find the area of this circular annulus. \\
\midrule
\textbf{Equation:} $x = 5*5*3.14 - 3*3*3.14 $ \\
\textbf{Answer:} \,\, 50.24 \\
\midrule
\textbf{With formula:} $x$ =  circle\_area(5) - circle\_area(3) \\
\bottomrule
\end{tabular}}
\caption{Example problem that requires geometry knowledge.}
\label{tb:mwp}
\end{center}
\end{table}

\begin{figure}[!h]
\centering
\includegraphics[width=0.8\linewidth]{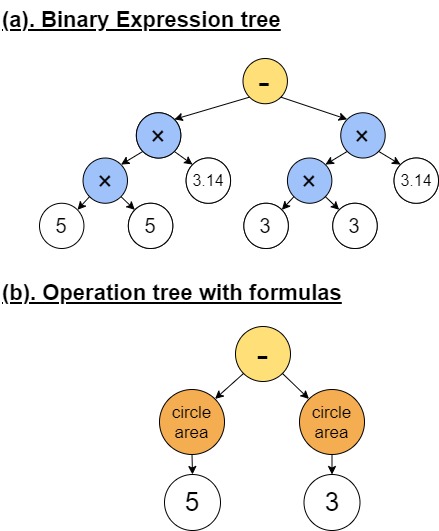}
\caption{(a). binary expression tree and (b). operation tree along with formulas for the problem in Table \ref{tb:mwp}.}
\label{fig:example}
\end{figure}

As with any QA task, solving an MWP requires the introduction of external knowledge or domain knowledge \cite{mishra2020towards}. However, current state-of-the-art solvers  \cite{xie2019goal,zhang2020graph,wu2020knowledge} do not address this issue explicitly. They learn to map the problem text into binary expression trees regardless of whether it requires any knowledge. This is counterintuitive for problems that need math concepts or formulas. As illustrated in Figure \ref{fig:example}(a), without explicitly using the corresponding area formula, the expression tree for the problem is lengthy and uninterpretable.

To address this issue, we propose a sequence-to-general tree (S2G) architecture where the nodes can be arbitrary math concepts or formulas with arbitrary number of arguments. In this way, our S2G model can learn to map the problem text into executable operation trees that contain different formulas across different domains. For example, S2G can learn to generate tree nodes that contain the required geometry formula for circles, as shown in Figure \ref{fig:example}(b), making the result more intuitive and explainable.

In addition, we propose a knowledge-guided mechanism to guide tree-decoding using a mathematical knowledge graph (KG). To evaluate our model, we also construct a middle-sized dataset consisting of 1,398 geometry word problems which require a diversified set of formulas.  Experimental results show that our S2G model can provide better performance and more interpretable results against strong baselines on problems that require domain knowledge.

The  main  contributions  of  this  paper are:
\begin{enumerate}
\item We propose a seq-to-general tree model that learns to map the problem text into operation trees where the nodes can be formulas with arbitrary number of arguments. This helps to incorporate domain knowledge into problem solving and produce interpretable results.
\item We design a knowledge-guided mechanism that guides tree decoding using mathematical knowledge graphs and GNNs.
\item We curate a middle-sized dataset that contains 1,398 geometry word problems. In addition, we annotate them with detailed formulas that can be readily converted into operation trees.
\end{enumerate}

\section{Seq2seq v.s. Seq2tree v.s. Seq2general}

\begin{figure}[htb]
\centering
\includegraphics[width=0.7\linewidth]{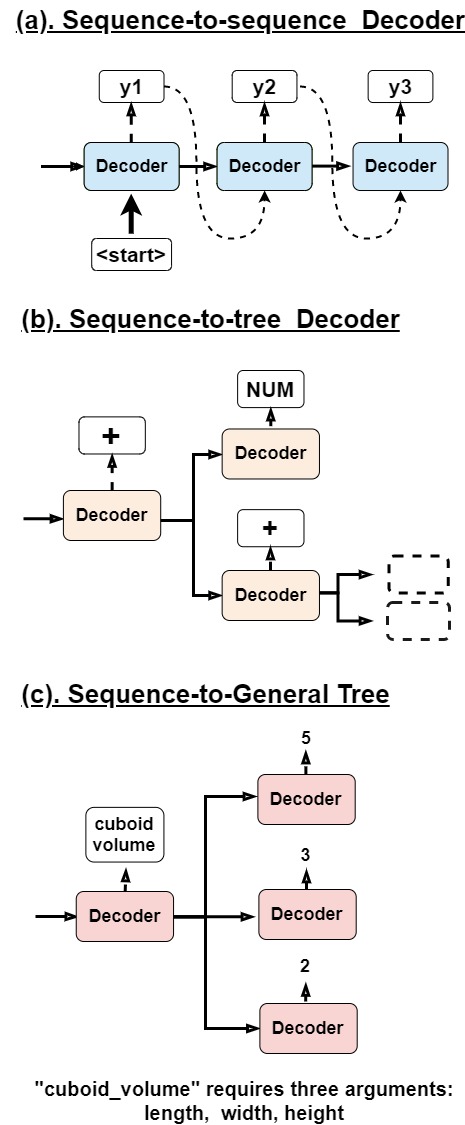}
\caption{Comparison between three types of decoding: (a) seq2seq, (b) seq2tree, and (c) seq-to-general tree.}
\label{fig:model_comparison}
\end{figure}

Our goal is to design a sequence-to-general tree model that learns to map the problem text into its corresponding operation tree. Before diving into the model, we first compare the decoding mechanisms between seq-to-seq, seq-to-tree and our seq-to-general tree solvers. Figure \ref{fig:model_comparison} illustrates the tree decoding process of these three types of model, respectively.

For seq2seq models, their decoder basically does two things: (1) predicting the current output and (2) generating the next state. These two steps can be conditioned on different information including the current state, the current input, or a context vector calculated using attention. The decoder would repeat these two steps until it outputs an end token.

For seq2tree models, however, this process is slightly different. The decoder predicts the current output as in seq2seq, but it will decide whether to generate the next state based on the current output. If the current output is a arithmetic operator, the decoder knows it should produce two child states, and these states are used to expand into its left and right children. If the current output is a number, then the decoder would end  the decoding process, so the current node becomes a leaf node. As a result, the whole decoding process resembles generating an expression tree in a top-down manner. 

In our work, we generalize the decoding process by making the decoder produce a variable number of children based on the type of the current output. If the output is a number or operator, the decoder would produce zero or two child states as before. If the output is a formula, the decoder will generate the pre-specified number of child states for this formula.

\section{Sequence-to-General Tree Model }
\label{sec:model}

In this section, we give a detailed description for each part of our S2G model.

\subsection{Encoder}

The main function of the encoder is to encode the problem text $P = (x_{1}, x_{2}, ..., x_{n})$ into a sequence of hidden states $(h_{1}, h_{2}, ..., h_{n})$ and their summary state $h_{encoder}$. The hidden states $h_{1}$ to $h_{n}$ are expected to contain the information for each input token $x_{1}$ to $x_{n}$, while the summary state $h_{encoder}$ is expected to capture the overall information of the problem.

Specifically, we use bidirectional gated recurrent units (GRU) \cite{cho2014learning} as our encoder. Given the current input $x_{t}$, the previous state $h_{t-1}$, and the next state $h_{t+1}$, the current state $h_{t} \in (h_{1}, h_{2}, ..., h_{n})$ can be calculated with: 
\begin{align}
&\overrightarrow{h_{t}}=\mathrm{GRU}(x_{t}, \overrightarrow{h_{t-1}}), \\ &\overleftarrow{h_{t}}=\mathrm{GRU}(x_{t}, \overleftarrow{h_{t+1}}), 
\label{eq:encoder1}
\end{align}
where the arrows represent different directions in the bidirectional encoding. After calculating the hidden state for each input token, we combine the last state of the forward and backward directions to get the summary state for the encoder: 

\begin{equation}
h_{encoder} =  \overleftarrow{h_{0}} + \overrightarrow{h_{n}}
\end{equation}

\subsection{Geometry Knowledge Graph}

To incorporate domain knowledge into problem solving, we propose to utilize the knowledge from mathematical knowledge graphs. The main idea is that given a formula predicted as the current node, we could use the physical meaning of its arguments to help us better predict its children. For example, if the current node is the formula for rectangle area, then we know its child nodes should be related to "length " and "width". We can thus use the node embeddings of "length" and "width" from a geometry KG to provide additional information for our solver.


We manually collect a geometry knowledge graph which contains the common geometry shapes (e.g., square, circle) and their geometry quantities (e.g., area, length), and we link these nodes to each other if they belong to the same shape.
To embed this KG, we employ a graph convolutional network (GCN) \cite{kipf2016semi} that transforms the KG into some vector space and calculates the embedding of each node. Given the feature matrix $X$ and the adjacency matrix $A$ of the KG, we use a two-layer GCN to encode it as follows:

\begin{equation}
Z = \mathrm{GCN}(X, A),
\end{equation}
where $Z = (z_{1}, ..., z_{n})$ are the node embeddings for each node in the graph. Then, we can use the embedding to represent the physical meaning of a certain formula argument in the decoding process.

\subsection{General Tree Decoder}

In the decoding stage, the decoder learns to produce the target operation trees in a recursive manner. It first predicts the current output $y_t$ in order to determine the number of children of the current node. Given the current decoder state $s_t$, the embedding of the last output $e_{(y_{t-1})}$, and the node embedding $z_t$ which represents the physical meaning in the knowledge graph, the probability of the current output $P(y_t)$ is calculated using:
\begin{align}
c_t &= \mathrm{Attention(e_{(y_{t-1})},h_1^n)} \\
z_t' &= \mathrm{Attention(z_t, h_1^n)} \\
P(y_t) &= \mathrm{Softmax}(W_y[s_t;e_{(y_{t-1})};c_t;z_t']),
\label{eq:start}
\end{align}
where $h_1^n$ is the encoder states $(h_{1}, ..., h_{n})$, $c_t$ is the context vector of  $e_{(y_{t-1})}$ with respect to $h_1^n$, and $z_t'$ is another context vector calculated using the node embedding $z_t$ and $h_1^n$. Specifically, we use additive attention \cite{bahdanau2014neural} to calculate these context vectors and use $h_{encoder}$ as the first decoder state $s_0$. Given the probability $P(y_t)$, we can then determine the output token $\hat{y_t}$: 
\begin{equation}
\hat{y_t} = \mathrm{argmax} P(y_t).
\end{equation}

Next, we predict the child states  conditioned on the required number of children for $\hat{y_t}$. Unlike previous binary-tree decoders that use two distinct DNNs to predict the left and right children respectively \cite{xie2019goal,zhang2020graph,wu2020knowledge}, we employ a GRU to predict a variable number of children. Given the current state $s_t$, its child states $s_{t_1}, ..., s_{t_n}$ are generated in a recurrent manner:
\begin{equation}
s_{t_i} = \mathrm{Decoder} (s_{t_{i-1}}; e_{(y_t)}; c_t),
\end{equation}
where we generate the first child $s_{t_{1}}$ using $s_t$, and the following child state $s_{t_i}$ using its previous sibling  $s_{t_{i-1}}$until we reach the required number of children. The decoder is basically a GRU followed by a linear projection layer and an activation function: 
\begin{align}
s_{t_i}'  &= \mathrm{GRU} ([e_{(y_t)}; c_t], s_{t_{i-1}}),\\
s_{t_i}  &= \mathrm{ReLU} (W_s s'_{t_i}),
\end{align}
where the input of GRU is the concatenation of $e_{(y_t)}$ and  $c_t$, $W_s$ is the linear projection layer, and ReLU is used as the activation function. After getting these child states, we push them into a stack and repeat the steps from Equation (5) to Equation (11) until all the states are realized into tokens.

\subsection{Training Objective}

For a problem and operation tree pair (\textit{P, T}), we follow previous seq2tree work \cite{xie2019goal,wu2020knowledge} and set our objective to minimize the negative log likelihood:
\begin{equation}
L(T,P) = \sum_{t=1}^n  - log P(y_t|s_t, P, KG).
\end{equation}

\section{Dataset}

To evaluate our S2G model on problems that require formulas, we curate a middle-sized dataset, \textit{GeometryQA}, that contains 1,398 geometry word problems. These problems are collected from Math23K \cite{wang2017deep} using the keywords of common geometric objects (e.g., circle, square, etc.) and their shapes (e.g., rectangular, circular, etc.). Then, we re-annotate each problem with their associated formulas if the problem belongs to one of the six major shapes: \textit{square}, \textit{cubic}, \textit{rectangle}, \textit{cuboid}, \textit{triangle} and \textit{circle}. Table \ref{tb:dataset} shows the overall statistics of GeometryQA and Table \ref{tb:formula} in Appendix \ref{appendix:dataset} shows the 11 formulas we used to annotate these problems. 

Note that not all problems in GeometryQA are annotated with formulas. About 16\% of the problems belong to other shapes (e.g., \textit{parallelogram}, \textit{rhombus}, etc.) which currently are not covered in our formula set. About 40\% are problems that contain geometric keywords but do not actually require any formulas. Table  \ref{tb:exception} shows such an example. We use these problems to test the robustness of our model. That is, S2G has to learn to apply the correct formulas or equations from misleading keywords (as shown in Table\ref{tb:exception}) and has to learn to generate both binary expression trees and operation trees as a whole.

\begin{table}[htb]
\begin{center}
\begin{tabular}{l l}
 \toprule
   \textbf{GeometryQA} & \\ 
 \midrule
   Number of problems & 1,398 \\
   Number of sentences/words & 5.4k / 41.1k \\
   Vocabulary size & 2,872 \\
 \midrule
   Annotated with formulas & 604 (43.20\%) \\
   Problems of other shapes & 225 (16.09\%) \\
   Formulas not required & 569 (40.70\%) \\
 \bottomrule
\end{tabular}
\caption{Dataset statistics of GeometryQA}
\label{tb:dataset}
\end{center}
\end{table}

\begin{table}[htb]
\begin{center}
\resizebox{\columnwidth}{!}{%
\begin{tabular}{l l}
 \toprule
\textbf{Problem:} The \underline{perimeter} of a \underline{rectangular} swim-\\
ming pool is 300 m. If you place a chair every \\
10 m all the way around its \underline{perimeter}, how many\\
chairs do you need? \\
\midrule
\textbf{Equation:} $x = 300 / 10$ \\
\textbf{Answer:} 30 \\
\bottomrule
\end{tabular}}
\caption{Example problem that contains misleading keywords (\underline{perimeter}, \underline{rectangular}) but do not require any geometry formulas.}
\label{tb:exception}
\end{center}
\end{table}

\section{Experiments}

\subsection{Implementation Details}

We implement our S2G model and the GNN module using Pytorch\footnote{https://pytorch.org/} and Pytorch Geometric\footnote{https://pytorch-geometric.readthedocs.io/en/latest/}. We set the dimension of word embedding to 128 and the dimension of the hidden state of GRU and GNN to 512. The  dropout  rate \cite{srivastava2014dropout}  is  set  to 0.5 and the batch size is 64. For optimization, we use ADAM \cite{kingma2014adam} with a learning rate of $10^{-3}$ and a weight decay of $10^{-5}$. Besides, we use a learning rate scheduler to reduce the learning rate by half every 20 epochs. During evaluation, we use beam search \cite{wiseman2016sequence} with a beam size of 5.

\subsection{Experimental Results on GeometryQA}

We evaluate our S2G model on GeometryQA to check whether it can learn to predict the corresponding operation tree for the geometry word problems. Table \ref{tb:exp2} shows the results of our S2G against other seq2tree SOTA models. S2G is trained using the re-annotated equations that contain formulas, while the baselines are trained using the original equations. 

First, we find that S2G has about 3.8\% performance gain over its baselines (with p-value $<$ 0.01). We attribute this to the fact that operation trees are easier to learn and generate since they are less lengthy and complex than binary expression trees. Hence, there is a better chance for S2G to produce the correct trees and arrive at the correct answers. 

Second, there is a small performance gain by adding Geometry KG. However, the improvement is not significant (with p-value$\approx$0.8).
We guess that is because the dataset currently has only six geometric objects, which is not complex enough to show the effectiveness of adding knowledge graphs.


\begin{table}[htb]
\begin{center}
\begin{tabular}{ l l}
 \toprule
   Model & Accuracy(\%) \\ 
 \midrule
KA-S2T \cite{wu2020knowledge} & 49.61\%\\
GTS \cite{xie2019goal} & 51.01\%  \\
\textbf{S2G} & \textbf{54.79\%} \\
\textbf{S2G + Geometry KG} & \textbf{54.99\%} \\
 \bottomrule
\end{tabular}
\caption{Answer accuracy of S2G and other SOTA seq2tree models on GeometryQA (all evaluated with 5-fold cross validation).}
\label{tb:exp2}
\end{center}
\end{table}

\section{Conclusion}

In this work, we proposed a sequence-to-general tree model (S2G) that aims to generalize previous seq2tree architectures. Our S2G can learn to generate executable operation trees where the nodes can be formulas with arbitrary number of arguments. By explicitly generating formulas as nodes, we make the predicted results more interpretable. Besides, we also proposed a knowledge-guided mechanism to guide the tree decoding using KGs and constructed a dataset in which problems are annotated with associated formulas. Experimental results showed that our S2G model can achieve better performance against strong baselines.



\bibliographystyle{acl_natbib}
\bibliography{acl2021}

\clearpage
\appendix

\section{Data Preprocessing}

In this section, we describe the data preprocessing steps required for our S2G model.

\subsection{Converting to prefix notation}

To perform top-down tree decoding, we follow \cite{xie2019goal} to convert the equations into their \textit{prefix notation}, where the operators are placed in front of their operands, rather than in between. In this way, the order of the equation tokens would match the order of decoding. In our case, we also need to consider the formulas used in the equation. For a formula in the form $"F(arg1, arg2)"$, we turn it into $"[F,  arg1, arg2]"$ so that it can fit into the prefix notation. Table \ref{tb:prefix} shows an example of this infix-to-prefix conversion for an equation with formulas. 

\begin{table}[htb]
\begin{center}
\begin{tabular}{l l}
 \toprule
\textbf{Problem:} The outer radius and inner radius of \\
a circular annulus are 5m and 3m respectively. \\
Find the area of this circular annulus.\\
\midrule
\textbf{Equation:} $x$ = circle\_area(5) - circle\_area(3) \\
\textbf{Prefix form:} [ -,  circle\_area, 5, circle\_area, 3] \\
\bottomrule
\end{tabular}
\caption{Infix-to-prefix conversion for an equation with formulas.}
\label{tb:prefix}
\end{center}
\end{table}

\subsection{Vocabulary}

We follow the canonical sequence-to-sequence architecture \cite{sutskever2014sequence} to prepare for the source vocabulary. For the target vocabulary, however, we have to take into consideration the way that humans solve MWPs. To solve a math problem, we use the numbers from the problem text (a dynamic vocabulary) and the mathematical operators learned before (a static vocabulary)  and try to compose them into an equation. Sometimes, we also need to use external constant numbers (a static vocabulary) that are not in the problem text but would appear in the equation (e.g., 1, 2,  or 3.14). These three types of vocabulary make up the vocabulary for the equations in arithmetic problems (equation \ref{eq:vocab1}).
\begin{equation}
V_{arith} = V_{number} \cup V_{op} \cup V_{const}
\label{eq:vocab1}
\end{equation}
We follow \cite{xie2019goal} to use a copy mechanism \cite{gu2016incorporating} to copy the numbers from the problem text. Hence, we can dynamically get the problem numbers during decoding. Besides, we extend the original vocabulary by adding domain-specific formulas into it so that the decoder can generate formulas during decoding (equation \ref{eq:vocab2}). Table \ref{tab:vocab} shows the overall vocabulary that we use for our decoder.
\begin{equation}
V_{target} = V_{number} \cup V_{op} \cup V_{const} \cup V_{formula}
\label{eq:vocab2}
\end{equation}
\begin{table}
\begin{center}
\begin{tabular}{ | c |  m{3.8cm} | } 
\hline
\textbf{Vocab Type} & \textbf{Instances} \\ 
\hline
Operator & +, -, *, /, \textasciicircum \\ 
\hline
Number & $\langle N0 \rangle$, $\langle N1 \rangle$, $\langle N2 \rangle$, ... \\ 
\hline
Constant & 1, 2, 3.14 \\ 
\hline
*Formula & circle\_area, square\_area, rectangle\_perimeter, and so on.  \\ 
\hline
\end{tabular}
\caption{Types of the vocabulary.}
\label{tab:vocab}
\end{center}
\end{table}

\section{GeometryQA}
\label{appendix:dataset}

Table \ref{tb:formula} shows the 11 formulas used for annotation.

\begin{table}[htb]
\begin{center}
\resizebox{\columnwidth}{!}{%
\begin{tabular}{l l c}
 \toprule
   Name & Formula & \# args\\ 
 \midrule
   \textbf{Square} &  \\
   square\_area & side * side  & 1 \\
   square\_perimeter & 4 * side  & 1 \\
 \midrule
   \textbf{Cubic} &  \\
   cubic\_volume & side*side*side  & 1 \\
 \midrule
  \textbf{Circle}\\
   circle\_area & $\pi$ * radius\textrm{\^{}}2  & 1 \\
   circumference\_r & 2 * $\pi$ * radius  & 1 \\
   circumference\_d & $\pi$ * diameter & 1 \\
 \midrule
  \textbf{Triangle}\\
   triangle\_area & base*height / 2 & 2 \\
  \midrule
 \textbf{Rectangle} \\
rectangle\_area & length * width & 2 \\
rectangle\_perimeter & 2 * (l+w) & 2 \\
  \midrule
 \textbf{Cuboid} \\
cuboid\_volume & l* w* height & 3 \\
cuboid\_surface & 2*(l*w+w*h+l*h) & 3 \\
\bottomrule
\end{tabular}}
\caption{Eleven geometry formulas used in annotating GeometryQA.
}
\label{tb:formula}
\end{center}
\end{table}

\section{Related Work}

In this section, we briefly introduce the progress of MWP solvers, and then we focus on topics that are closer to our work, including seq2tree solvers and knowledge graphs for problem solving.

\subsection{Math Word Problem Solving}

Ever since 1960s, efforts have
been made to build automatic math word problem solving systems \cite{feigenbaum1963computers,bobrow1964natural}. Statistical solvers learn to map problem features into corresponding equation templates or operations to solve the problem \cite{kushman2014learning,hosseini2014learning,mitra2016learning,liang-etal-2016-meaning,liang-etal-2018-meaning,roy2018mapping}. For example, \citet{kushman2014learning} propose  to align MWPs to their templates, while \citet{hosseini2014learning} propose to find the operations by verb categorization. Semantic parsing approaches, on the other hand, parse the problem into intermediate representations using semantic parsers \cite{shi2015automatically,koncel2015parsing,huang2017learning}.

Recently, neural architectures have emerged as a dominant paradigm in math word problem solving. \citet{wang2017deep} first attempt to use a seq2seq solver that utilize encoder-decoder architectures to encode the problem text and then decode into equations in a way similar to machine translation. Copy mechanism \cite{huang2018neural} or attention mechanisms \cite{li2019modeling} are introduced to improvement the performance of seq2seq models. These seq2seq models, however, suffer from producing invalid equations, like a binary operator with three operands, because there is no grammatical constraint on its sequential decoding. To solve this problem, seq2tree models are proposed to bring into the grammatical constraints \cite{xie2019goal,liu2019tree}. We will give a more detailed introduction to seq2tree models in Section \ref{sec:s2t}.

\subsection{Sequence-to-Tree Models}
\label{sec:s2t}

To convert text into structured representations, several research strands have utilized sequence-to-tree models. \citet{dong2016language} first use seq2tree on semantic parsing to translate text into structured logical forms. Similar frameworks are also adopted for code generation \cite{yin2017syntactic,rabinovich2017abstract} where they translate code snippets into executable representations or abstract syntax trees (ASTs).

Inspired by their ideas, MWP solving also adopts seq2tree to map the problem text into expression trees. This introduces a constraint that the non-leaf nodes of the tree should be operators and leaf nodes be numbers, and thus the resulted equations are always guaranteed to be valid. Most seq2tree solvers choose bidirectional LSTM or GRU as their text encoder and use two separate models to predict the left and right nodes during decoding respectively    \cite{xie2019goal,zhang2020graph,wu2020knowledge,li-etal-2020-graph}. Our model differs from the previous that we use a single RNN-based decoder to predict a variable number of children nodes during decoding. In addition, our model can predict formulas as nodes that increase the interpretability of the model outputs, while previous solvers can only predict basic arithmetic operators.

\subsection{Knowledge Graph for Math Word Problem Solving}
\label{related-kg}

To incorporate external knowledge into problem solving, some solvers utilize graph convolutional networks \cite{kipf2016semi} or graph attention networks \cite{velivckovic2017graph} to encode knowledge graphs (KGs) as an additional input. \citet{wu2020knowledge} proposed to incorporate commonsense knowledge from external knowledge bases. They constructed a dynamic KG for each problem to model the relationship between the entities in the problem. For example, "daisy" and "rose" would be linked to their category "flower" so that the solver can use this hyperonymy information when counting the number of flowers. \citet{zhang2020graph} proposed to build graphs that model the quantity-related information using dependency parsing and POS tagging tools \cite{manning2014stanford}. Their graphs provide better quantity representations to the solver. Our model differs from previous models that we aim to incorporate domain knowledge from mathematical KGs rather than from  commonsense knowledge bases.

\end{document}